\documentclass[letterpaper]{article}
\usepackage{natbib,alifeconf}

\title{More Bang For Your Buck: Quorum-Sensing Capabilities Improve\\ the Efficacy of Suicidal Altruism}
\author{Anya Elaine Johnson$^{1,5}$, Eli Strauss$^{2,5}$, Rodney Pickett$^{1,5}$, Christoph Adami$^{3,5}$, Ian Dworkin$^{2,5}$, \& Heather J. Goldsby$^{4,5}$ \\
\\
$^1$Computer Science and Engineering, Michigan State University, East Lansing, MI, USA \\
$^2$Zoology, Michigan State University, East Lansing, MI, USA \\
$^3$Microbiology and Molecular Genetics / Physics and Astronomy, Michigan State University, East Lansing, MI, USA \\
$^4$Biology, University of Washington, Seattle, WA, USA, 
$^5$BEACON Center for the Study of Evolution in Action \\
anyaejo@msu.edu}

\begin{document}
\maketitle

\begin{abstract}
Within the context of evolution, an altruistic act that benefits the receiving individual at the expense of the acting individual is a puzzling phenomenon. An extreme form of altruism can be found in colicinogenic \emph{E. coli}. These suicidal altruists explode, releasing colicins that kill unrelated individuals, which are not colicin resistant. By committing suicide, the altruist makes it more likely that its kin will have less competition. The benefits of this strategy rely on the number of competitors and kin nearby. If the organism explodes at an inopportune time, the suicidal act may not harm any competitors. Communication could enable organisms to act altruistically when environmental conditions suggest that that strategy would be most beneficial. Quorum sensing is a form of communication in which bacteria produce a protein and gauge the amount of that protein around them. Quorum sensing is one means by which bacteria sense the biotic factors around them and determine when to produce products, such as antibiotics, that influence competition. Suicidal altruists could use quorum sensing to determine when exploding is most beneficial, but it is challenging to study the selective forces at work in microbes. To address these challenges, we use digital evolution (a form of experimental evolution that uses self-replicating computer programs as organisms) to investigate the effects of enabling altruistic organisms to communicate via quorum sensing. We found that quorum-sensing altruists killed a greater number of competitors per explosion, winning competitions against non-communicative altruists. These findings indicate that quorum sensing could increase the beneficial effect of altruism and the suite of conditions under which it will evolve.
\end{abstract}

\section{Introduction}

An organism behaves altruistically when it performs an action that lowers its own fitness in order to increase the fitness of another organism \citep{WEST:2007is}. Altruistic organisms ranging from bacteria to humans are found frequently in nature \citep{Chao:1981wj, Bowles:2006jl}. We might expect that any altruistic genes should be selected against because the altruistic organism will have a lower fitness than a non-altruistic organism benefitting from an altruistic act (a ``cheater"). A particularly challenging form of altruism to explain is suicidal altruism, where the altruistic organism dies to increase the fitness of other organisms  \citep{Velicer:2000tm,Khare:2009hy}. However, inclusive fitness theory and multi-level selection theory both describe conditions under which altruism--even suicidal altruism--can evolve \citep{Traulsen:2006iu,Hamilton:1963wp}. According to inclusive fitness theory, altruism is favored by selection when an organism's altruistic action directly benefits its kin, who are likely to share the same altruistic gene(s)  \citep{Khare:2009hy}. However, an open question is how do altruistic organisms know when to behave altruistically? For example, how do altruistic organisms know when their kin surround them and would benefit from their actions? 

%Cite also Fletcher & Zwick. J Theor Biol. 2004 Jun 7;228(3):303-13.
% and "Unifying the Theories of Inclusive Fitness and Reciprocal Altruism" vol. 168, no. 2 the american naturalist august 2006 — CA (maybe)

Under certain conditions, colicinogenic \emph{E. coli} commit suicide to kill nearby competitors and thus free resources for their kin \citep{Chao:1981wj}. Specifically, these \emph{E. coli} stochastically produce a toxin until they explode. Once an \emph{E. coli} explodes, the toxin is released, killing all surrounding organisms that are not resistant. Because the exploding organism's kin are usually resistant to the toxin, they survive the explosion and benefit from the decrease in surrounding competition for resources \citep{Chao:1981wj}. However, the prevailing environmental conditions, such as the number of competitors and kin surrounding the altruistic organism, determine the benefits of the suicidal action. Specifically, when an organism explodes, it is possible that either competitors are not present (and thus the explosion does not affect the competition for resources) or kin are not nearby (and thus are unable to take advantage of the accessible resources). In both of these cases, the altruistic organism committed suicide without providing any benefit to its kin. 

An organism that can communicate with constituents to gain information about its environment could potentially improve decisions regarding when to perform altruistic actions. Quorum sensing is a type of communication found in many species of bacteria \citep{Diggle:2007ul,Davies:1998fq}, whereby the bacteria can assess how many constituent organisms surround them. Quorum sensing involves an organism releasing a small amount of a signaling protein and gauging the amount of that same protein that has been released by other bacteria around it \citep{Bassler:2002in}. A number of organisms combine quorum sensing with antibiotic production to strategically kill competitors \citep{Chandler:2012ja}. In these situations, producing an antibiotic is an expensive action and requires the cooperation of other constituents to produce enough to harm competitors. If too few organisms are producing the antibiotic, the action is expensive and does not accrue any benefits. However, if the organisms achieve quorum and ensure that enough constituents are also producing the antibiotic, then it is likely they can harm or kill the competitor. Thus, quorum sensing is used to enable organisms to determine whether the environmental conditions are favorable to perform expensive and possibly altruistic acts.  

In this study, we explore whether altruistic organisms that do not have access to information about their environment would benefit from quorum-sensing capabilities. While enabling colicinogenic \emph{E. coli} to use quorum sensing would determine the benefit of such a strategy, the challenges involved with performing that experiment are numerous. They include the difficulty associated with evolving complex traits, such as quorum sensing and suicidal altruism, and the generation time of \emph{E. coli}. To address these experimental challenges, we use Avida, a digital evolution system \citep{Ofria:2004bf}. Within Avida, computer programs (``digital organisms'') self-replicate with a chance of mutation and compete for space within their environment. Avida has been used to study evolutionary topics such as division of labor \citep{Goldsby:2012tz} and the evolution of biological complexity \citep{Lenski:2003vy}. Additionally, Avida has been used to study suicidal altruism \citep{Goings:2004wv} and quorum sensing \citep{Beckmann:2009ko} in isolation, making it the ideal platform for evolution experiments that combine these two complex traits.  In this study, we use digital organisms that explode based on stochastic factors and quorum-sensing information when available. When an organism explodes, it kills competitors, and thus behaves similarly to colicinogenic \emph{E. coli}.  We study whether such organisms will evolve to use quorum-sensing capabilities and whether these capabilities provide a competitive advantage against altruists who cannot communicate. 

 %Adami1998 is my book "Introduction to Artificial Life"

\section{Related Work}

Suicidal altruism and communication have been studied extensively in both organic and digital systems \citep{Berngruber:2013ja, Bordereau:1997en, Chao:1981wj,Crespi:2001fy, Davies:1998fq, Diggle:2007ul, Foster:2006ts, Goings:2004wv, Goldsby:2012tz, Hamilton:1964wz, Hamilton:1963wp, Kerr:2004gx}. Using both types of systems, researchers have found strong evidence that the benefits from higher inclusive fitness enables the success of suicidal-altruism strategies \citep{Chao:1981wj, Goings:2004wv, Berngruber:2013ja, Bordereau:1997en,Crespi:2001fy, Foster:2006ts}. Furthermore, studies in both organic and digital systems have found that cooperation mediated by quorum sensing can be a successful strategy in a number of environments \citep{Diggle:2007ul, Davies:1998fq}. Here we discuss relevant studies surrounding the evolution of suicidal altruism and quorum sensing within organic and digital systems.

Within nature, suicidal altruistic acts are generally performed to increase the success of the altruistic organism's kin \citep{Chao:1981wj,Berngruber:2013ja}. For example, colicinogenic \emph{E. coli} stochastically kill competitors in an explosion of toxins, thus the surviving colony members have less competition for resources \citep{Chao:1981wj}. Additionally, in some species of \emph{E. coli}, when a bacterium of \emph{E. coli} carrying the suicide gene \emph{Lit} is infected by a lytic phage, it may kill itself to prevent the spread of the pathogen to the rest of the colony \citep{Berngruber:2013ja}. 

There are a number of organisms that use quorum sensing to regulate altruistic behavior \citep{Miller:2001uo,Dworkin:1985hy, Chandler:2012ja}. For example, \emph{Myxococus xanthus} is a bacterium that uses quorum sensing to detect cell density. When confronted with starvation, at high cell density this bacterium forms fruiting bodies \citep{Miller:2001uo}. For the fruiting body to be successful, the majority of organisms must lyse to form the stalk and only a small fraction of the organisms reproduce as spores \citep{Dworkin:1985hy}.  Additionally, a number of species of bacteria use quorum sensing to kill competitors with antibiotics \citep{Chandler:2012ja}. Competitors only die once the amount of antibiotics in the environment reaches a certain concentration \citep{Chandler:2012ja}. If the antibiotic that is produced is too diluted, competitors may become resistant to it. As such, the bacteria are most likely to succeed in harming the competitors if a group of them simultaneously produce the antibiotic. Quorum sensing is used to establish that such a group is present and available to produce the antibiotic.

Altruism and quorum sensing have previously been studied in digital systems as well. Goings \emph{et al.} studied suicidal altruism modeled on the colicinogenic \emph{E. coli's} behavior \citep{Goings:2004wv}. They discovered that suicidal altruism could evolve in the Avida digital evolution platform, even though the organisms could not communicate with each other. Suicidal altruism was most likely to evolve in a structured environment, where an organism had a high probability of being surrounded by its kin, rather than in a well-mixed environment, where an organism's kin were distributed randomly. In a subsequent study investigating environmental factors that influence the evolution of altruism, we found that large population sizes and high mutation rates increased the evolution of suicidal altruism \citep{Johnson:prep}. Additionally, Beckmann \emph{et al.} studied the evolution of quorum sensing in Avida \citep{Beckmann:2009ko}. They determined that digital organisms could evolve to use quorum sensing to suppress replication when population density reached a specific threshold. Furthermore, they established that organisms using quorum sensing could utilize it within  populations up to 400 times larger than those in which the organisms had evolved. While these studies explored the conditions under which suicidal altruism and quorum sensing arise independently, here we investigate whether the conjunction of these traits is beneficial.

\section{Methods}

\begin{figure}[t]
\begin{center}
\includegraphics[width=3in, angle=0]{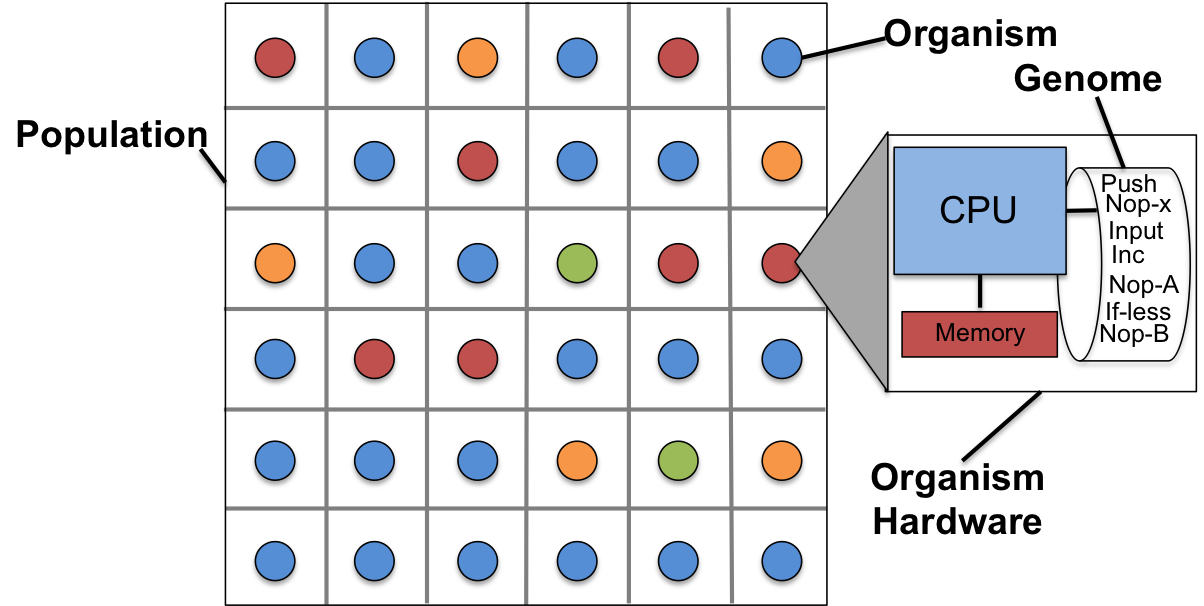}
\vskip 0.25cm
\vspace{-4ex}
\caption{{\bf An Avida population.} Each square represents a cell potentially occupied by an organism. Each circle represents an organism, where color denotes the genotype. Each organism consists of a genome, a virtual CPU for executing instructions, and memory space for storing values.
}
\label{population}
\end{center}
\vspace{-2ex}
\end{figure}

For these experiments, we used the Avida digital evolution platform \citep{Ofria:2004bf}. Within Avida, digital organisms live on a toroidal grid where they compete for space (Figure~\ref{population}). A digital organism consists of a computer program (its genome) and virtual hardware upon which the program is executed. Included in this virtual hardware are three registers that the organism can use to store and manipulate numbers. Each genome consists of a series of instructions that enable the organism to input numbers from and output numbers to the environment (IO), manipulate numbers, and self-replicate.  Each organism occupies a specific cell in the grid. When an organism replicates, it creates two daughter organisms that inherit a potentially-mutated version of their parent's genome. One daughter organism replaces the parent and the other daughter organism is placed into a surrounding cell, potentially killing the current occupant. 

\subsection*{Altruism and Quorum Sensing Instructions}
For this study, in addition to the standard set of Avida instructions \citep{Ofria:2004bf}, we developed several instructions that enable organisms to evolve to use quorum sensing and altruistically explode (potentially killing competitors).   

If evolved, the \emph{quorum-sense} instruction enables the digital organism to sense how many related organisms surround it. For these experiments, we consider two organisms to be \emph{related} if their genomes differ by three or fewer instructions.  An organism's neighborhood consists of any organisms in the 5x5 grid surrounding the organism. When an organism executes the \emph{quorum-sense} instruction, the following computation occurs: (1) The proportion of neighboring cells filled by related organisms is computed. (2) If this proportion is \emph{less} than a \emph{quorum-threshold} value specified by the organism, then one of the organism's registers is set to TRUE.  While the organisms can use this capability in isolation, we envision that they will use it to assess whether environmental conditions are suitable for altruistic explosions. 

\begin{figure}[t]
\begin{center}
\includegraphics[width=3in, angle=0]{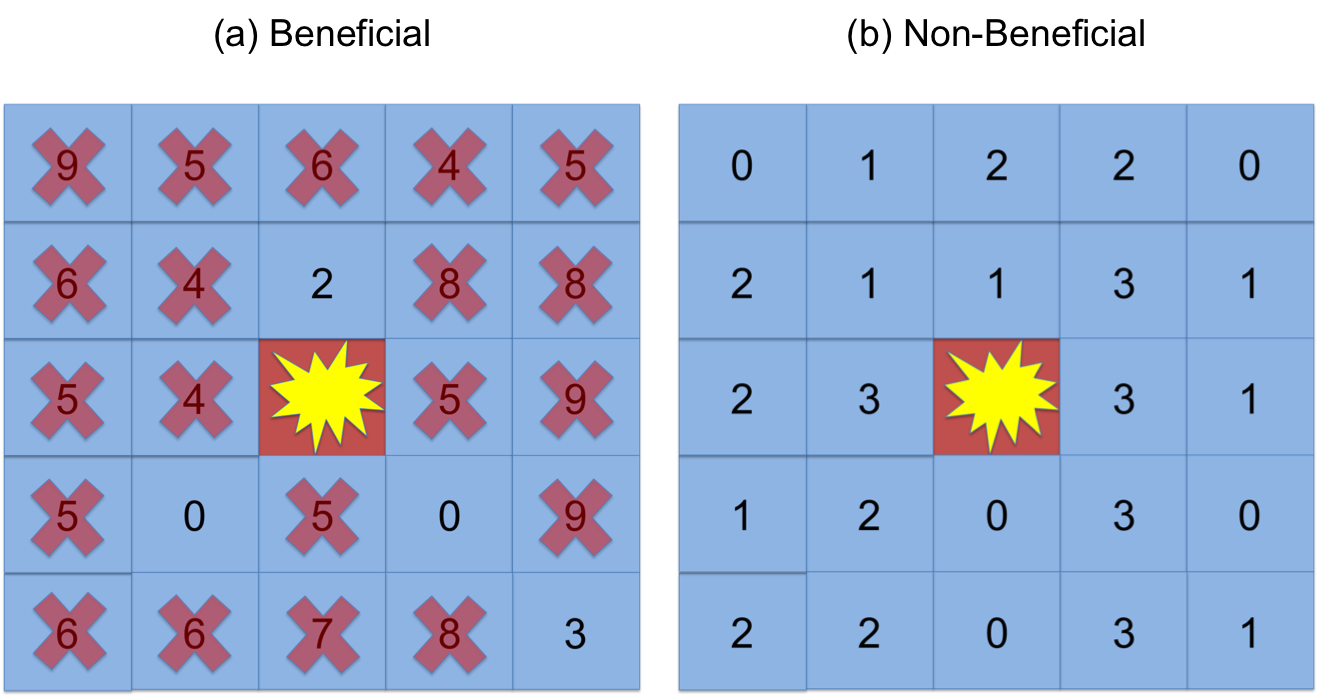}
\vskip 0.25cm
\vspace{-4ex}
\caption{{\bf The results of an organism exploding in a beneficial and non-beneficial environment.} Each 5x5 grid represents a subsection of the Avida world, where each square represents an organism. Within the grid, the center square represents the focal, exploding organism. The numbers specify the number of genomic differences from the focal organism. Squares containing an X represent organisms that are killed by the explosion because they are unrelated to the focal organism (i.e., they have four or more genomic differences). (a) This grid depicts the effects of a beneficial explosion. Many of the organisms surrounding the exploding organism are unrelated and are killed by the toxin released by the exploding organism.  (b) This grid depicts the effects of a non-beneficial explosion. All the organisms in the exploding organism's neighborhood are related, thus the focal organism's explosion did not kill any competitors.
}
\label{explodecomp}
\end{center}
\vspace{-2ex}
\end{figure}

\begin{figure}[t]
\begin{center}
\includegraphics[width=3in, angle=0]{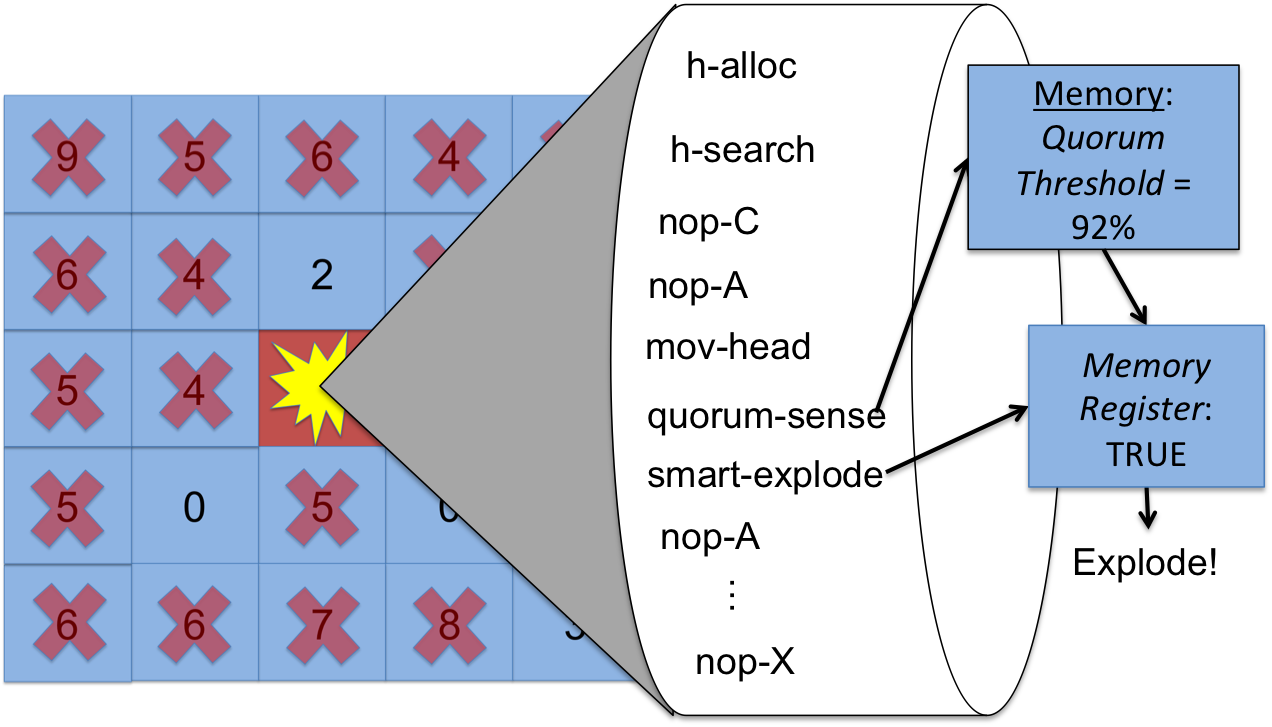}
\vskip 0.25cm
\vspace{-4ex}
\caption{{\bf An organism executing the \emph{quorum-sense} and \emph{smart-explode} instructions.}  A portion of the genome and internal state of the exploding organism (exploding center square) are shown. When the organism executes the \emph{quorum-sense} instruction, a register in the organism's memory is set to TRUE because the percentage of related individuals is less than 92\% (the organism's quorum threshold). The organism then executes the \emph{smart-explode} instruction, which checks the same register in memory and, because it is TRUE, explodes with a 5\% chance. 
}
\label{explodegenome}
\end{center}
\vspace{-2ex}
\end{figure}

The \emph{smart-explode} instruction, if evolved, enables an organism to use quorum information in determining whether or not to explode. In many examples within nature, quorum sensing is used as a trigger for actions (e.g., a sufficient number of bacteria are present to produce an antibiotic). However, for this study, we envision that quorum sensing will be used to inhibit explosions under unsuitable conditions (Figure~\ref{explodecomp}), though the population must still evolve to use the instruction in this way. Specifically, the presence of a quorum indicates that the focal organism is surrounded by related organisms and an explosion would not free up resources. As such, when the \emph{smart-explode} instruction is executed, the following computation occurs (Figure~\ref{explodegenome}): If the register specified by the organism (by default this is the same register that \emph{quorum-sense} sets) is set to TRUE, the organism explodes with 5\% probability.  Otherwise, the instruction has no effect. The probability of exploding was 5\% to be congruent with previous digital evolution studies \citep{Goings:2004wv}. 

Several factors influence whether organisms successfully make use of these instructions to sense their environment and react accordingly. First, organisms are not able to directly sense the presence of competitors, but must infer this information from the proportion of related organisms around them. When an organism detects that only a few kin are in its neighborhood, it cannot be certain that the rest of its neighborhood contains competitors. Some or all of the non-kin cells could be empty spaces instead, rendering an explosion useless. This instruction mimics the quorum-sensing capabilities of natural organisms, who only sense the presence of conspecifics. Additionally, the register used by the \emph{smart-explode} instruction can also be used by other instructions. An organism could have a suicidal-altruism strategy that does not use quorum sensing. The \emph{quorum-sense} and \emph{smart-explode} instructions provide the most accurate information when executed sequentially. This ensures that the organism is using the information gathered by the \emph{quorum-sense} instruction and that that information reflects the current status of the surrounding environment. For example, if an organism executes the \emph{quorum-sense} instruction early within its lifetime, but waits to execute the \emph{smart-explode} instruction until substantially later, then the quorum-sense information may no longer be accurate. This could result in an explosion that would not otherwise have occurred.  

\subsection*{Experimental Parameters}
For each treatment, we configured an environment in which organisms evolved on a 60x60 toroidal grid for 30,000 updates. An update is the time unit used by Avida. One update allows an organism to execute 30 instructions on average and 10 updates is approximately one generation. All experiments had a mutation rate of 0.02 genomic. These are standard parameters employed by Avida experiments \citep{Goings:2004wv, Goldsby:2012tz}.

We performed two main types of experiments: single-lineage evolution and competition assays. The single-lineage evolution experiments started with one ancestor with a genome of length 100. The ancestor contained only instructions necessary for replication and no-operation (nop) instructions. In the quorum-sensing-altruists treatment of this experiment, the standard set of instructions and the \emph{quorum-sense} and \emph{smart-explode} instructions were available via mutation. In the non-quorum-sensing-altruists treatment (a control), the standard set of instructions and the \emph{explode} instruction were available via mutation. When an organism executes the \emph{explode} instruction, it  explodes with a 5\% probability.

To assess whether the \emph{quorum-sense} and \emph{smart-explode} instructions were undergoing positive selection, we compared how frequently they were executed to the frequency with which the \emph{nop-Y} instruction was executed. \emph{Nop-Y} is a no-operation instruction that does not affect the operation of the organism and therefore was used as a control. 

We performed several competition assays with distinct starting proportions of the different lineages. For these experiments, the organisms were not able to mutate the \emph{quorum-sense}, \emph{smart-explode} or \emph{explode} instructions in or out of their genome, which enabled us to assess the competitive capability of these lineages. The first set of competition assays started with a colony of non-quorum-sensing altruists and a colony of quorum-sensing altruists in equal proportions.   The second set of experiments began with 95\% non-quorum-sensing altruists and 5\% quorum-sensing altruists. This permutation of initial conditions enabled us to test whether quorum-sensing altruists could invade a colony of  non-quorum-sensing altruists. Finally, we determined if quorum-sensing altruists could repel non-quorum-sensing altruists by configuring the initial population to contain 5\% non-quorum-sensing altruists and 95\% quorum-sensing altruists.  

Configuration files and data for all experiments are available at https://github.com/anyaejohnson/Communication14.

\section*{Results and Discussion}
In this study, we first explore whether organisms will evolve to use quorum sensing to inform their decisions about when to behave altruistically. Next, we examine the strategies organisms evolved to determine how many related organisms constitute a quorum. Finally, we determine if organisms that can use quorum sensing to inform their decisions regarding if and when to altruistically commit suicide have a competitive advantage over organisms that do not have access to quorum-sensing information.

\subsection*{Can organisms evolve to use quorum information to decide when to commit suicide?}
For our first experiment, we investigated whether quorum-sensing and suicidal-altruism behaviors would evolve, rise in frequency, and ultimately stabilize in a population. To assess whether these capabilities were beneficial, we evolved a colony of organisms from an ancestor organism that could only reproduce, but quorum-sensing and suicidal-altruism instructions were available via mutation.

\begin{figure}[t]
\begin{center}
\includegraphics[width=2.5in, angle=0]{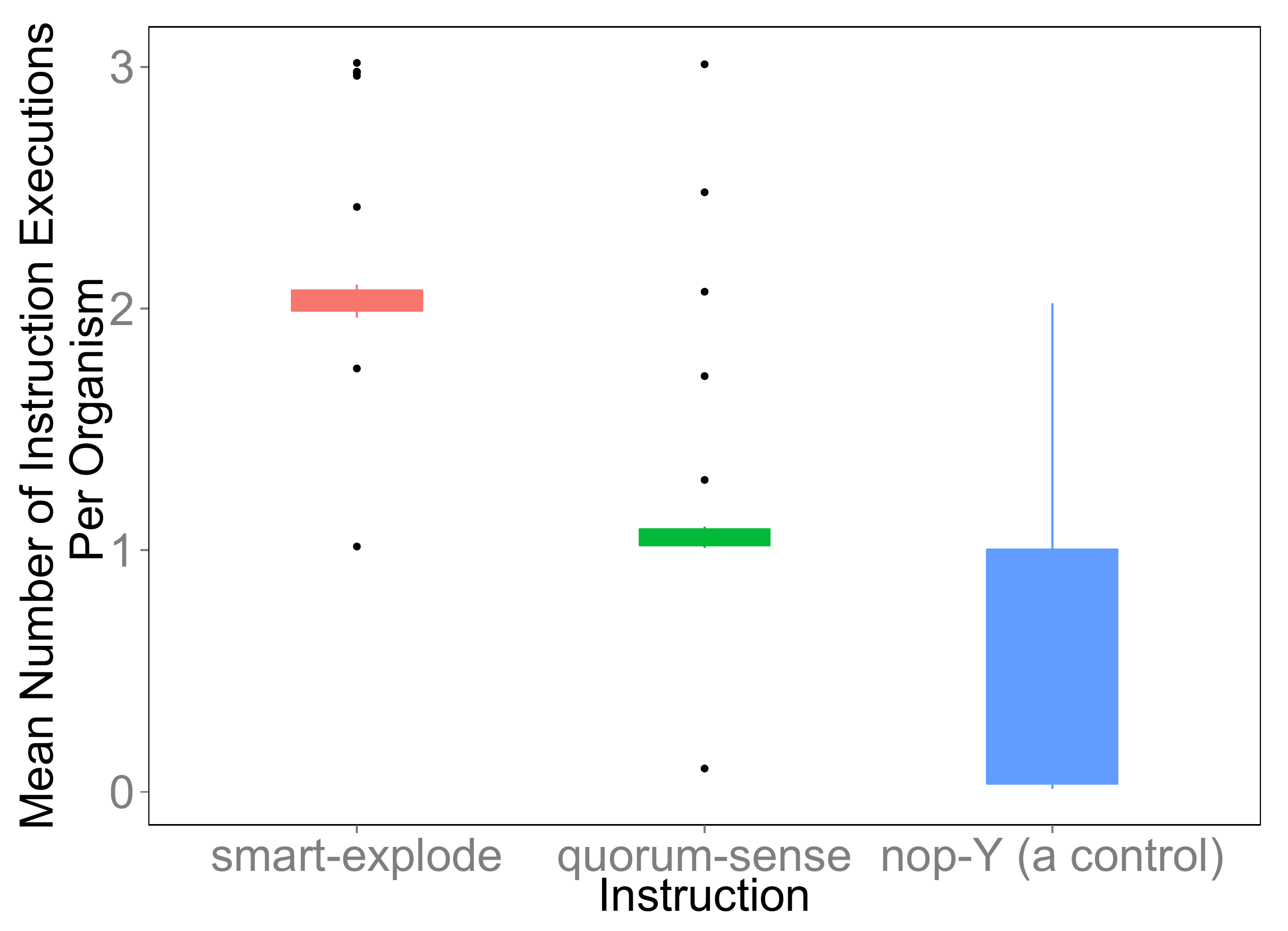}
\vskip 0.25cm
\vspace{-4ex}
\caption{{\bf The average instruction executions for smart-explode, quorum-sense and a neutral instruction per organism in the population.}  The \emph{smart-explode} instruction was used 1.95 more per organism than the neutral instruction (\emph{nop-Y}) in the last 100 updates. The \emph{quorum-sense} instruction was used 0.99 more per organism than the neutral instruction (\emph{nop-Y}) in the last 100 updates (95\% CI for difference between \emph{smart-explode} and neutral instruction is 1.92 to 1.97 executions per organism, Wilcox rank sum test $p < 0.0001$; 95\% CI difference \emph{quorum-sense} instruction executions and neutral instruction is 0.96 to 1.00 executions per organism, Wilcox rank sum test $p < 0.0001$, error bars represent 95\% confidence intervals). Both quorum sensing and suicidal altruism are under positive selection in this environment.
}
\label{num_inst_experimental}
\end{center}
\vspace{-2ex}
\end{figure}

 Figure~\ref{num_inst_experimental} shows the execution rates per organism of the quorum-sensing, suicidal-altruism, and neutral instructions. In agreement with previous work \citep{Goings:2004wv}, suicidal altruism was used an average of 1.95 time more frequently than a neutral instruction per organism by the final time point (median number suicidal-altruism instruction executions per organism = 2.00, neutral instruction executions per organism = 0.04, 95\% confidence interval of difference is 1.92 to 1.97 instruction executions,  Wilcox rank sum test $p < 0.0001$). Organisms also evolved to use the quorum-sensing instruction an average of 0.99 more frequently than the neutral instruction per organism by the final time point (median number of quorum-sensing executions per organism = 1.03, neutral instruction executions = 0.04, 95\% confidence interval of difference is 0.96 to 1.00 more instruction executions per organism, Wilcox rank sum test $p < 0.0001$). These results demonstrate that both suicidal-altruism and quorum-sensing capabilities were selected for. 

\subsection*{How many related organisms constitute a quorum?}
Quorum-sensing altruists are required to set a threshold to determine how many organisms constitute a quorum. This quorum threshold also implicitly sets the number of potential competitors or empty spaces that must be present for an explosion to be considered beneficial. For our quorum-sensing-altruist treatment, we explored how many organisms constituted a quorum by recording the value of the quorum threshold each time an organism exploded (once the exploding mechanism evolved into the population). 

\begin{figure}[t]
\begin{center}
\includegraphics[width=2.0in, angle=0]{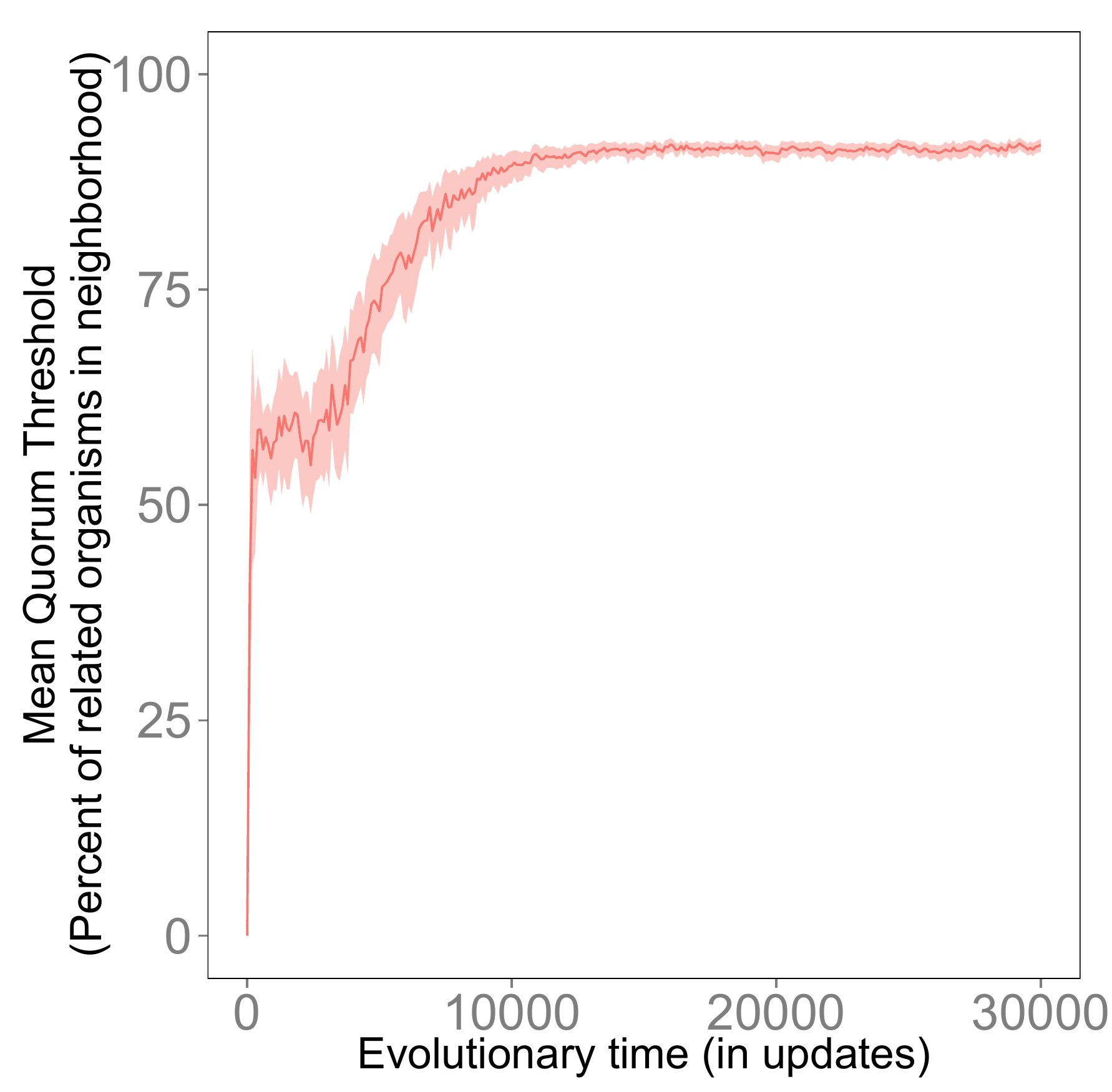}
\vskip 0.25cm
\vspace{-4ex}
\caption{{\bf Mean quorum threshold evolved by quorum-sensing altruists.} The quorum-sensing altruists evolved to use a \emph{quorum threshold} of approximately 92\% related organisms in their neighborhood. Within the context of this study, it is likely that an organism using quorum sensing would explode if its neighborhood contained 8\% or more unrelated organisms or empty spaces. 
}
\label{threshold_nonnoisy}
\end{center}
\vspace{-2ex}
\end{figure}

Figure~\ref{threshold_nonnoisy} shows that the organisms evolved a median quorum threshold that corresponded to approximately 92\% of the organisms around the focal organism being related (95\% confidence interval is 91.29\% to 92.63\%). This percentage indicates that organisms using the quorum-sensing instruction with suicidal altruism would only probabilistically explode if one or more unrelated organisms or empty spaces were around the organism. This mechanism prevents the organism from exploding when the organism is completely surrounded by related organisms.

\subsection*{Is quorum information beneficial?}
Ideally, an organism should use quorum sensing to assess whether its environmental conditions are such that an explosion would benefit its kin. However, the act of quorum sensing also has a cost: the organism must spend time gathering information about its environment. Within our study, this cost is experienced by the organism as the CPU cycles it devotes to gathering and using quorum-sensing information. To assess whether the costs of using quorum-sensing information outweigh the benefits of additional environmental information, we compared the behavior of quorum-sensing altruists to non-quorum-sensing altruists. 

\begin{figure}[t]
\begin{center}
\includegraphics[width=2.8in, angle=0]{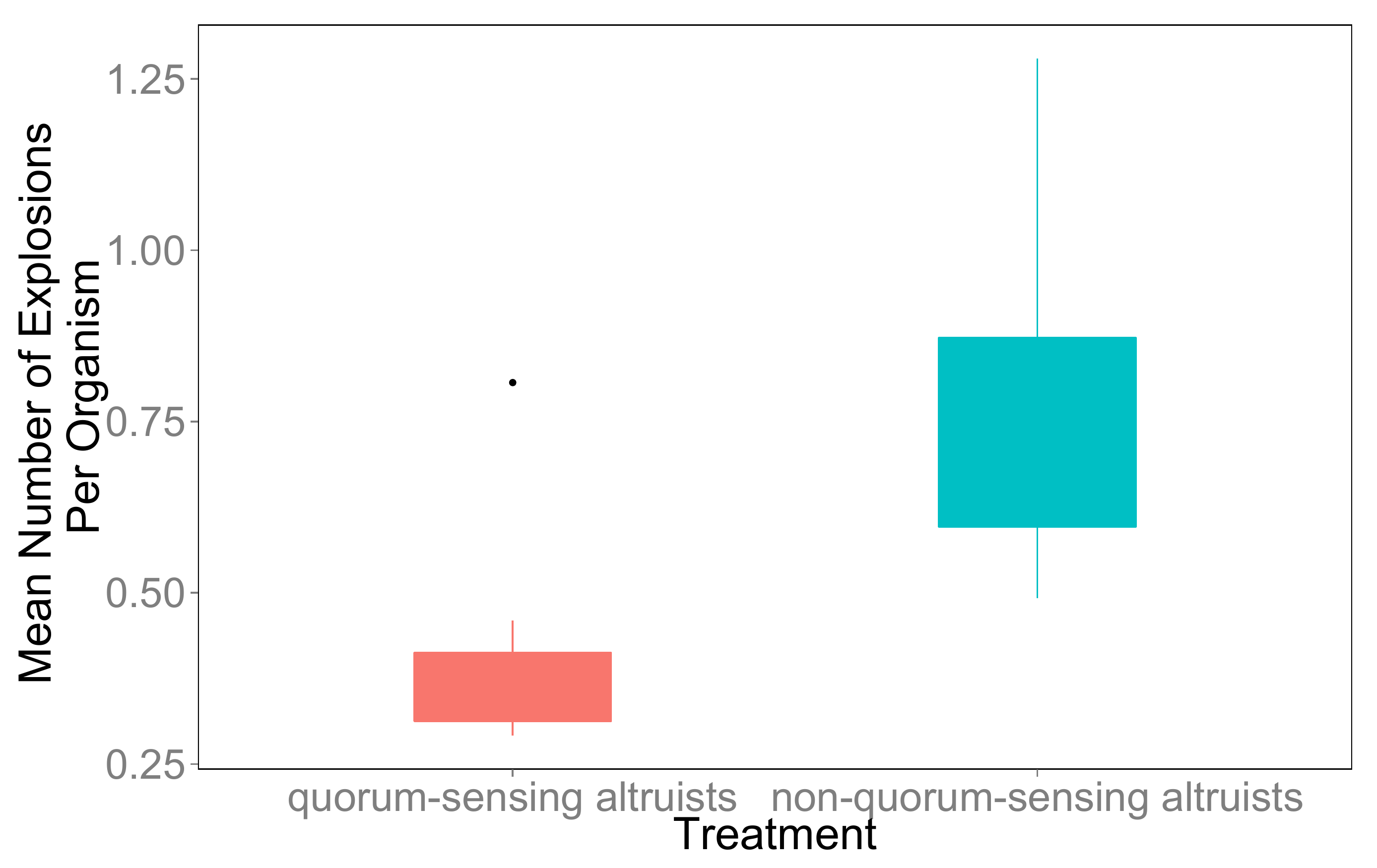}
\vskip 0.25cm
\vspace{-4ex}
\caption{{\bf Mean number of explosions per organism for the quorum-sensing altruists and the non-quorum-sensing altruists.} At the final time point in the treatments, the quorum-sensing altruists explode 0.35 less per organism on average than the non-quorum-sensing altruists (median quorum-sensing-altruist explosions = 0.35, median non-quorum-sensing-altruist explosions = 0.75, 95\% CI for difference is 0.28 to 0.45 explosions per organism, Wilcox rank sum test $p < 0.0001$). 
}
\label{explosionscomp}
\end{center}
\vspace{-2ex}
\end{figure}

To investigate how these costs and benefits influence altruistic behavior, we determined the number of explosions per organism that occurred within the quorum-sensing-altruist and non-quorum-sensing-altruist treatments.  Figure~\ref{explosionscomp} shows the number of explosions per organism in each treatment. The quorum-sensing altruists explode an average of 0.35 times less frequently per organism than the non-quorum-sensing altruists (median quorum-sensing altruist explosions = 0.35, median non-quorum-sensing-altruist explosions = 0.75, 95\% confidence interval of difference is 0.28 to 0.45 fewer explosions per organism, Wilcox rank sum test $p < 0.0001$).  These results may indicate that quorum sensing allows the quorum-sensing altruists to avoid exploding in a detrimental environment or they may indicate that quorum sensing is not adaptive. 

%0.7  more is confusing. I think this is an absolute number again and not a ratio. But the way you write this, people expect a ratio, and when saying "x times more" you expect x >1. So now the reader is really confused [CA].  

\begin{figure}[t]
\begin{center}
\includegraphics[width=2.2in, angle=0]{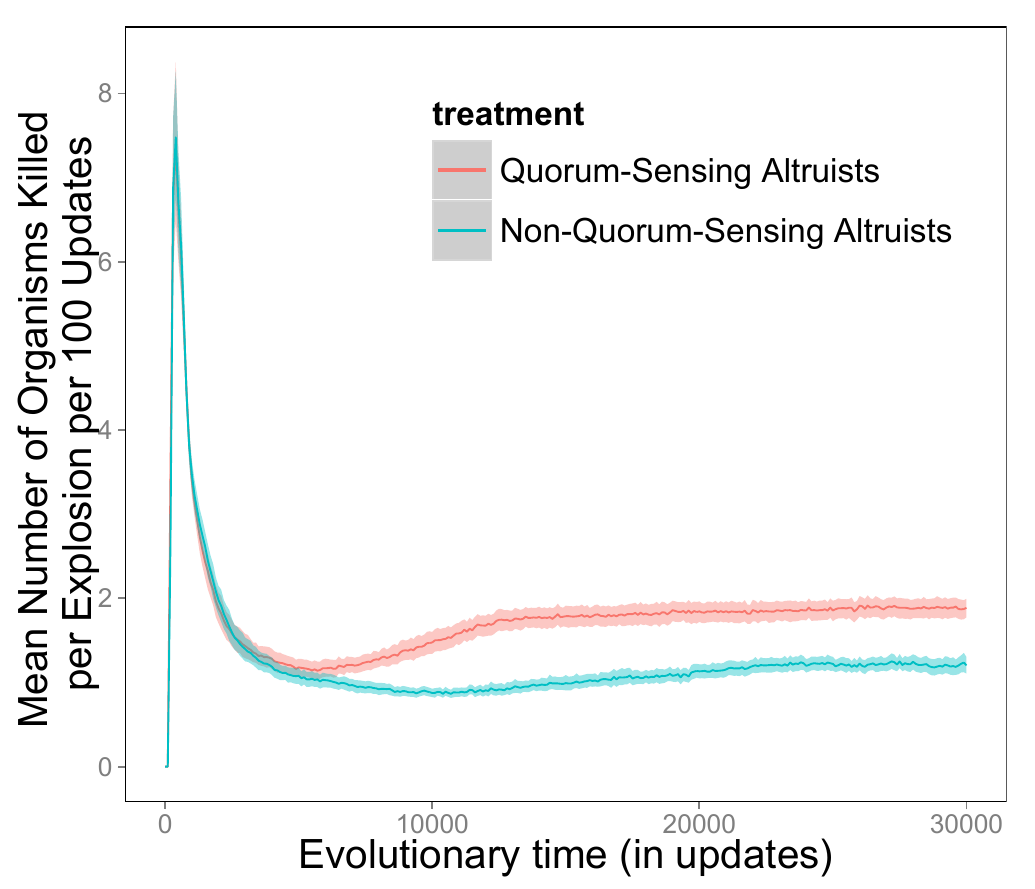}
\vskip 0.25cm
\vspace{-4ex}
\caption{{\bf A comparison of the effects of quorum-sensing and non-quorum-sensing-altruist explosions.} Shown is the average number of organisms killed per explosion every 100 updates. The quorum-sensing-altruist treatment is more efficient, killing 0.702 more organisms on average than the non-quorum-sensing altruists (mean number of organisms killed per explosion by quorum-sensing altruists = 2.004, mean number of organisms killed per explosion by non-quorum-sensing altruists = 1.231, 95\% CI of difference is 0.567 to 0.847, Wilcox rank sum test $p < 0.0001$). We also calculated the average number of organisms killed per explosion normalized by population size. When normalized by population size, the average number of organisms killed per explosion by quorum-sensing altruists is 0.000197 more organisms than the non-quorum-sensing altruists (mean number of organisms killed per explosion normalized by population size for quorum-sensing altruists = 0.00058, non-quorum-sensing altruists = 0.00037, 95\% CI of difference is 0.000161 to 0.000243, Wilcox rank sum test $p < 0.0001$). 
}
\label{avg_orgs}
\end{center}
\vspace{-2ex}
\end{figure}

To assess whether quorum-sensing altruists are deriving more benefit from each explosion than non-quorum-sensing altruists, we compared the mean number of organisms killed per explosion in each treatment. Figure~\ref{avg_orgs} shows that the mean number of organisms killed per explosion by the quorum-sensing altruists and non-quorum-sensing altruists is 2.004 and 1.231, respectively. Thus, on average, the quorum-sensing altruists kill 0.702 more organisms per explosion than by the non-quorum-sensing altruists (95\% confidence interval of difference is 0.567 to 0.847 organisms killed per explosion, Wilcox rank sum test $p < 0.0001$). The population size is also larger in the quorum-sensing-altruists treatment, which could be contributing to the greater number of organisms killed (median population size in the final time point of quorum-sensing-altruists treatment = 3423.5, non-quorum-sensing-altruists treatment = 3371.5). When the mean number of organisms killed is normalized by population size, the quorum-sensing altruists kill an average of 0.000197 more organisms per explosion than the non-quorum-sensing altruists (mean number of organisms killed per explosion normalized by population size for quorum-sensing altruists = 0.00058, for non-quorum-sensing altruists = 0.00037, 95\% confidence interval of difference is 0.000161 to 0.000243, Wilcox rank sum test $p < 0.0001$). Therefore, even when larger population size is accounted for, quorum-sensing altruists kill more competing organisms than the non-quorum-sensing altruists on average. These results suggest that the benefits of quorum sensing may outweigh the costs.

\subsection*{Does quorum sensing provide a competitive advantage for suicidal altruists?}
We have demonstrated that quorum-sensing altruists kill more organisms per explosion than non-quorum-sensing altruists. However, these results do not yet demonstrate that quorum sensing provides a competitive advantage. While quorum-sensing altruists may kill more organisms per explosion, the costs of quorum sensing may yet outweigh this benefit. Therefore, we performed competition assays in which quorum-sensing altruists compete directly against non-quorum-sensing altruists. Our competition assays used two lineages of organisms. One lineage was required to use the quorum-sensing-altruist strategy and the other lineage was required to use the non-quorum-sensing-altruist strategy. To focus on the competitive capabilities of these two strategies, we hand wrote both ancestor organisms and prevented the quorum-sensing and suicidal-altruism instructions from being accessible via mutation.  

\begin{figure*}[t]
\begin{center}
\includegraphics[width=6in, angle=0]{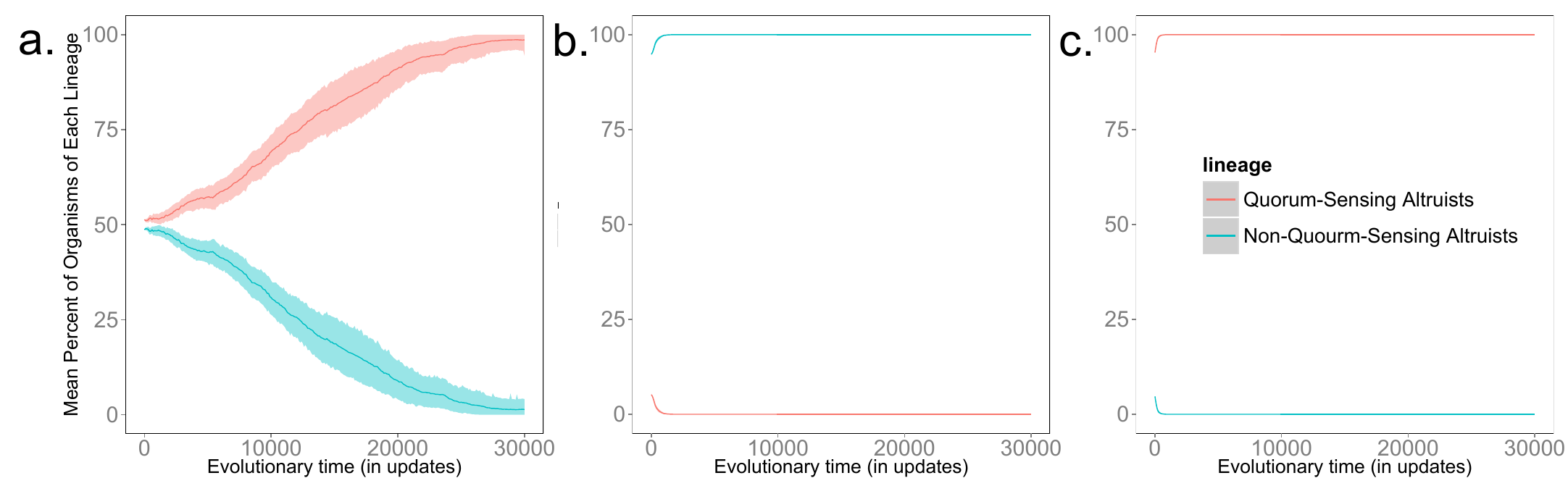}
\vskip 0.25cm
\vspace{-4ex}
\caption{{\bf Mean percent of quorum-sensing altruists (red) and non-quorum-sensing altruists (blue) over time.} In (a), each lineage started at 50\% of the population and then evolved in competition. In (b), the quorum-sensing altruists started at 5\%  of the population and the non-quorum-sensing altruists started at 95\%. In (c), the quorum-sensing altruists started at 95\% of the population and the non-quorum-sensing altruists started at 5\%. The quorum-sensing altruists achieved fixation by the end of the experiment when starting at equal proportion or the majority proportion. The non-quorum-sensing altruists achieved fixation when they started in the majority.
}
\label{comp_all}
\end{center}
\vspace{-2ex}
\end{figure*}

For our first competition assay, we filled half the population with copies of the quorum-sensing-altruist ancestor and the other half with copies of the non-quorum-sensing-altruist ancestor. Figure~\ref{comp_all}a shows the mean percent of organisms of each lineage over time.  Out of 30 replicates, the quorum-sensing lineage fixed 30 times. Therefore, despite the additional costs of quorum sensing, the information gathered through quorum sensing provides a competitive advantage.  

There are many situations in nature where a colony of organisms already exists. Therefore, we examined whether or not the quorum-sensing altruists could invade an existing population of non-quorum-sensing altruists.  Specifically, we performed competition assays in which the lineage of quorum-sensing altruists started at only 5\% of the population and the non-quorum-sensing altruists started at 95\% of the population. As seen in Figure~\ref{comp_all}b, the quorum-sensing altruists were not able to invade a pre-established population of the non-quorum-sensing altruists and went extinct in 30/30 replicates. 

Finally, we examined the reverse situation and explored whether quorum-sensing altruists could repel invading non-quorum-sensing altruists. Figure~\ref{comp_all}c depicts our results. In this case, the quorum-sensing altruists are able to repel the non-quorum-sensing altruists in all 30 replicates. These experiments demonstrate that quorum-sensing altruists have a competitive advantage over non-quorum-sensing altruists when they start at an equal or greater proportion of the population. 

\section*{Conclusions}

We explored whether organisms would evolve to use quorum sensing to inform a suicidal altruism strategy and whether such a strategy would have a competitive advantage. We found that when we enable digital organisms to set their own \emph{quorum threshold}, they evolved a threshold of 92.1\% on average, meaning that organisms evolved to require at least one competitor or empty space to be in their neighborhood in order to explode. Additionally, we found that quorum sensing increased the benefit of suicidal altruism (i.e., the number of competitors killed per explosion significantly increased). Finally, we found that quorum-sensing altruists outcompeted non-quorum-sensing altruists when starting at equal proportions or in the majority. Future work could explore intermediate values to address whether quorum-sensing altruists could invade a pre-existing population. In further experiments, we also explored how the accuracy of the quorum-sensing mechanisms affected their use~\citep{MSU-CSE-14-2}. In general, quorum-sensing accuracy was correlated with an advantage in competitions against non-quorum-sensing altruists. 

Suicidal altruism is a critical aspect of survival strategies in many natural systems, despite its high cost \citep{Chao:1981wj, Shorter:2011hb}. Communication has often been implicated as the element that might aid the evolution of altruism \citep{WEST:2007is, Turner:1999jg, Strassmann:2011hk}. This study offers support that: (1) Quorum-sensing capabilities can be used to make altruistic acts more effective. (2) Communicating altruists can outcompete altruists that are not able to communicate. In this study we purposefully kept the cost of quorum-sensing low to facilitate evolution. Future work could explore how increasing this cost affects the competitive advantage of quorum-sensing altruists. 

\section*{Acknowledgments}
We thank Charles Ofria, David Bryson, and the Devolab for their insightful contributions to this work. This research was supported by funding from the BEACON Center for the Study of Evolution in Action and the NSF. This material is based in part upon work supported by the National Science Foundation under Cooperative Agreement No. DBI-0939454 and No. OCI-1122620. Any opinions, findings, and conclusions or recommendations expressed in this material are those of the author(s) and do not necessarily reflect the views of the National Science Foundation.

\footnotesize
\bibliographystyle{apalike}
\bibliography{references}

%\begin{table}[h]
%\center{
%\begin{tabular}{|c|c|c|c|}\hline
%Name & Result & Bonus $b_i$ & Difficulty\\ \hline\hline
%Echo & I/O   & 1 & --\\
%Not  & $\neg A$ & 2 & 1 \\
%Nand & $\neg(A\wedge B)$ & 2 & 1 \\
%Not Or & $\neg A \vee B$ & 3 & 2 \\
%And  &  $ A \wedge B $   & 3 & 2 \\
%Or   &  $ A \vee B $     & 4 & 3 \\
%And Not & $A\wedge\neg B$& 4 & 3 \\
%Nor  & $\neg(A\vee B)$   & 5 & 4 \\
%Xor  & $ A\ {\rm xor}\ B$ &   6 & 4 \\
%Equals &$\neg(A\ {\rm xor}\ B)$&6& 4 \\ \hline
%\end{tabular}
%}
%\vskip 0.25cm
%\caption{Logical calculations on random inputs $A$ and $B$ rewarded,
%bonuses, and difficulty (in minimum number of {\tt nand} instructions
%required). Bonuses $b_i$ increase the speed of a CPU by a factor
%$\nu_i=1+2^{b_i-3}$.}
%\end{table}

\end{document}